\documentclass{article}

\usepackage[dblblindworkshop,final]{neurips_2026} 
\workshoptitle{AXIOM: Foundations of Efficient Deep Learning}

\usepackage[utf8]{inputenc}
\usepackage[T1]{fontenc}    
\usepackage{hyperref}       
\usepackage{url}            
\usepackage{booktabs}       
\usepackage{amsfonts}       
\usepackage{nicefrac}       
\usepackage{microtype}      
\usepackage{xcolor}         

\usepackage{amsthm}

\newtheorem{theorem}{Theorem}
\newtheorem{remark}{Remark}

\newtheorem{definition}{Definition}

\newcommand{\bbR}{\mathbb{R}}

\newcommand{\1}[0]{\mathbf{1}}

\newcommand{\calK}{\mathcal{K}}

\newcommand{\calO}{\mathcal{O}}

\newif\ifmynotes
\mynotesfalse

\newcommand{\inputDim}{m}
\newcommand{\inputDimANN}{m^\text{ANN}}
\newcommand{\inputDimSNN}{m^\text{SNN}}
\newcommand{\outputDim}{n}
\newcommand{\width}{w}
\newcommand{\depth}{L}
\newcommand{\depthANN}{L^\text{ANN}}
\newcommand{\depthSNN}{L^\text{SNN}}
\newcommand{\depthRNN}{L^\text{RNN}}
\newcommand{\sparsity}{S}

\newcommand{\regions}{\calK}
\newcommand{\regionsANN}{\calK_\text{ANN}}
\newcommand{\regionsSNN}{\calK_\text{SNN}}
\newcommand{\regionsRNN}{\calK_\text{RNN}}

\newcommand{\nnMult}{\text{$\boldsymbol{\otimes}$}_\text{NN}}
\newcommand{\nnAdd}{\text{$\boldsymbol{\oplus}$}_\text{NN}}
\newcommand{\nnAct}{\text{A}_\text{NN}}
\newcommand{\nnMem}{\text{M}_\text{NN}}
\newcommand{\nnMemOne}{\text{M}_\text{NN}^1}

\newcommand{\annMult}{\text{$\boldsymbol{\otimes}$}_\text{ANN}}
\newcommand{\annAdd}{\text{$\boldsymbol{\oplus}$}_\text{ANN}}
\newcommand{\annAct}{\text{A}_\text{ANN}}
\newcommand{\annMem}{\text{M}_\text{ANN}}

\newcommand{\rnnMult}{\text{$\boldsymbol{\otimes}$}_\text{RNN}}
\newcommand{\rnnAdd}{\text{$\boldsymbol{\oplus}$}_\text{RNN}}
\newcommand{\rnnAct}{\text{A}_\text{RNN}}
\newcommand{\rnnMem}{\text{M}_\text{RNN}}

\newcommand{\snnMult}{\text{$\boldsymbol{\otimes}$}_\text{SNN}}
\newcommand{\snnAdd}{\text{$\boldsymbol{\oplus}$}_\text{SNN}}
\newcommand{\snnAct}{\text{A}_\text{SNN}}
\newcommand{\snnMem}{\text{M}_\text{SNN}}
\newcommand{\snnMemOne}{\text{M}_\text{SNN}^1}

\newcommand{\energyNN}{\text{E}_\text{NN}}
\newcommand{\energyAnn}{\text{E}_\text{ANN}}
\newcommand{\energyRnn}{\text{E}_\text{RNN}}
\newcommand{\energySnn}{\text{E}_\text{SNN}}

\newcommand{\energyOp}{\text{E}_\text{op}}
\newcommand{\nnOp}{\text{op}_\text{NN}}

\newcommand{\energyMult}{\text{E}_\text{$\otimes$}}
\newcommand{\energyAdd}{\text{E}_\text{$\oplus$}}
\newcommand{\energyAct}{\text{E}_\text{A}}
\newcommand{\energyMem}{\text{E}_\text{M}}
\newcommand{\energyMemOne}{\text{E}_\text{M}^1}

\newcommand{\energyMac}{\text{E}_\text{mac}}
\newcommand{\energyAcc}{\text{E}_\text{ac}}
\newcommand{\energyDiff}{\text{E}_\text{diff}^\text{ANN}}

\newcommand{\inputDimRNN}{m^\text{RNN}}

\newcommand{\energyDiffR}{\text{E}_\text{diff}^\text{RNN}}

\usepackage{graphicx}
\usepackage{xcolor}

\usepackage{amsmath}
\usepackage{amsfonts}

\usepackage{cleveref}
\usepackage{wrapfig}

\crefname{section}{Sec.}{Sec.}
\crefname{subsection}{Sec.}{Sec.}
\crefname{figure}{Fig.}{Fig.}
\crefname{algorithm}{Alg.}{Alg.}
\crefname{table}{Tab.}{Tab.}
\crefname{example}{Ex.}{Ex.}
\crefname{definition}{Def.}{Def.}
\crefname{proposition}{Prop.}{Prop.}
\crefname{corollary}{Cor.}{Cor.}
\crefname{theorem}{Thm.}{Thm.}
\crefname{lemma}{Lemma}{Lemmas}
\crefname{appendix}{Appendix}{Appendix}
\title{Towards an Expressivity-Normalized Energy-Demand Comparison of ANNs, RNNs, and SNNs}

\author{
 Miriam Kranzlmüller$^{1, 2}$
    \quad
Pascal Esser$^{1, 2}$
  \quad
  \textbf{Gitta Kutyniok}$^{1, 2, 3, 4}$\\
    $^1$Department of Mathematics, LMU Munich\\
    $^2$Munich Center for Machine Learning (MCML)\\
    $^3$Institute for Robotics and Mechatronics, DLR-German Aerospace Center  \\
    $^4$Department of Physics and Technology, University of Tromsø  \\
}

\begin{document}

\workshoptitle{AXIOM}
\maketitle

\begin{abstract}
Spiking neural networks (SNNs) are often regarded as energy-efficient alternatives to conventional architectures, yet their advantage depends on both network architecture and data properties. We develop an analytical framework to compare fully-connected ReLU ANNs, Elman RNNs, and integrate-and-fire SNNs for time-series data with respect to an expressivity-normalized measure of theoretical energy efficiency. By relating an inference-energy model to theoretical bounds on representational expressivity, we derive an expressivity-normalized efficiency ratio and explicit thresholds in network width, spike sparsity, and network depth. Our analysis characterizes the regimes in which event-driven computation offsets the temporal overhead of SNNs, providing capacity-aware principles for designing energy-efficient temporal networks. It shows that ANNs and RNNs exceed SNNs in expressivity-normalized efficiency only in specific regimes.
\end{abstract}

\section{Introduction}
The inference cost of modern neural networks has become a central obstacle to their use in resource-constrained settings. Conventional artificial neural networks (ANNs) and recurrent neural networks (RNNs) rely on dense matrix operations, which result in substantial computational costs. In recent years, spiking neural networks (SNNs) have emerged as an alternative in which neurons communicate through discrete events and update their state only when driven by incoming spikes \cite{gerstner2002spiking,maass1997networks}. Their sparse, event-driven nature makes SNNs particularly suitable for temporal and spatio-temporal data \cite{dominguez2018deep,ibad2022hyperparameter,lv2024efficient,matenczuk2021financ}.

However, the energy advantage commonly attributed to SNNs is not without constraints. Existing analytical comparisons often examine the number of multiplication and addition operations required by ANNs and SNNs \cite{barchid2023spiking,kim2022beyond,lemaire2022synaptic}. 
Yet arithmetic operations alone do not determine inference cost. Memory access and data movement may dominate the energy budget \cite{hoefflin2025hardware,horowitz2014computing}, while the sequential processing of an input over multiple timesteps can introduce substantial repeated computation and communication. Accordingly, hardware measurements may differ significantly from simplified operation-count estimates \cite{bhattacharjee2024are,zhouer2022efficiency,thienbutr2024energy,xu2025eventbased}.
Therefore, establishing a fair comparison baseline is crucial for accurately quantifying the energy efficiency of SNNs relative to ANNs and RNNs. Existing energy evaluations predominantly enforce architectural parity by holding parameter counts and network sizes constant across the compared network types \cite{lemaire2023analytical, lemaire2022synaptic, yan2025reconsidering} or assess energy consumption via conversion methods, where a trained ANN is converted into an SNN \cite{andrei2024deep, deng2020rethinking, sengupta2018going}.
While these approaches offer insights under fixed structures, they do not account for differences in model expressivity. 
In a time-series setting, an ANN can process a temporally concatenated input in a single forward pass, whereas SNNs and RNNs process the same sequence sequentially. Differences in width, depth, and temporal processing may therefore change both energy expenditure and the ability of the architectures to partition the input space. To compare their energy requirements on a common theoretical scale, we use the number of induced linear or constant regions as a proxy for representational capacity \cite{montufar2014number,nguyen2025time,serra2018bounding}. 
Discrete-time SNNs with leaky integrate-and-fire neurons, fixed threshold, and reset by subtraction can be modeled as recurrent networks with binary activations \cite{neftci2019surrogate}. With respect to region counts, an RNN unrolled over $T$ timesteps is a weight-tied feedforward network, so the bounds of ANNs apply in principle but do not exploit the weight sharing, while existing empirical comparisons between spiking and recurrent networks match accuracy rather than capacity \cite{yin2021accurate}.

In this work, we develop a theoretical framework evaluating the energy demand of fully-connected ReLU ANNs, Elman RNNs, and integrate-and-fire SNNs with respect to an expressivity-normalized measure of theoretical energy efficiency. We introduce a hardware-informed inference-energy model that accounts for arithmetic operations as well as multi- and single-bit memory accesses. Combining this model with upper bounds on the number of regions realized by the three architectures, we define an expressivity-normalized energy-efficiency ratio. We then derive explicit thresholds for network width, spike sparsity, and depth for the ANN and RNN that characterize when the SNN is more energy-efficient at comparable theoretical expressivity. Our analysis shows that \emph{SNN efficiency is inherently conditional: sparse event-driven computation can offset temporal overhead in specific architectural and sparsity regimes.}

We provide network definitions in appendix \ref{app:neural_networks} and their number of operations and memory accesses in appendix \ref{app:energy_derivation}. All derivations and proofs are provided in appendix \ref{app:energy_expressivity}.
\section{A Hardware-Informed Energy Model}
\textbf{Network Definitions.} 
We consider a fully-connected feedforward ANN and an Elman RNN, both consisting of $\depth+1$ layers and employing ReLU activation functions. 
Here, $\depth$ denotes the \emph{depth}, the number of hidden layers, and $(\inputDim,n_1,\dots,n_\depth, \outputDim)$ are the layer \emph{widths}. We choose a fixed hidden layer width $w = n_\ell$ for $\ell \in [L]$.
The input and output dimensions are $\inputDim$ and $\outputDim$.
We compare them to an integrate-and-fire SNN \cite{maass1997networks}.
While feedforward ANNs execute computations in a single forward pass, SNN neurons preserve an internal state, the \emph{membrane potential} $u_\ell(t)$, integrating the signals over discrete timesteps $t \in [T]$ \cite{nguyen2025time}. 
They spike as soon as their membrane potential exceeds a threshold~$\theta_\ell$.
In detail, we use an SNN whose output is derived from the membrane potential at the final timestep $T$, $y = u_{L+1}(T)$.
To enable a comparison, we choose the following relation between the input dimensions:
$\inputDim^{\text{ANN}}=\inputDim^{\text{SNN}} T$ and $\inputDim^{\text{RNN}}=\inputDim^{\text{SNN}}$.
While the ANN processes the time series as a single tensor, capturing temporal dependencies in a single timestep with a higher input-to-hidden ratio, the RNN and SNN process one timestep at a time, operating sequentially.

\textbf{Energy Demands.} Quantifying the energy consumption of a neural network during inference is challenging due to its strong dependence on specific hardware implementations. 
To circumvent the limitations of direct hardware-level measurements, we establish an arithmetical metric that enables an estimation of the energy demand based on model characteristics.
In general, the energy consumption of a neural network is determined by its architecture, specifically its depth $\depth$, width $\width$, and input/output dimensions $\inputDim$, $\outputDim$, as well as the data-dependent sequence length $T$ and the activation sparsity $S$, representing the share of neurons with output zero. 
Here, the sparsity $S$ of an SNN indicates the fraction of non-spiking neurons, averaged over all timesteps.
Note that, in theory, the sparsity of ANNs and RNNs can also be considered when examining certain implementations of matrix multiplications. 
However, we assume dense multiplications in ANNs and RNNs.
We model the total energy demand, $\energyNN$, as the sum of the energy costs for operations and memory accesses
\begin{equation}
\label{eq:total_energy}
    \energyNN(\inputDim, \outputDim, \width, \depth, S, T) = \energyMult \cdot \nnMult + \energyAdd \cdot \nnAdd 
    + \energyAct \cdot \nnAct + \energyMem \cdot \nnMem + \energyMemOne \cdot \nnMemOne\:. 
\end{equation}
Here, $\energyOp$ denotes the energy cost per specific operation (multiplication $\otimes$, addition $\oplus$, activation A,  multi-bit M, or single-bit $\text{M}^1$ memory access), while $\nnOp$ represents the total count of the corresponding operations executed during inference.
Based on \cite{horowitz2014computing} and normalized to a 32-bit addition, we assume the following simplified relationships between the operations:
$
    \energyMult = 4\energyAdd,\, \energyAct = \energyAdd, \,\energyMem = 5\energyAdd, \,\energyMemOne=0.2\energyAdd
$.
Memory access accounts for the largest portion of computational energy consumption. Although single-bit accesses reduce the overhead of data transmission, our bit-proportional energy scaling represents an idealized approximation rather than the exact physics of semiconductors.
The energy demand is calculated based on \eqref{eq:total_energy} for all three networks, where the energy for multiply-and-accumulate operations is $\energyMac = \energyMult + \energyAdd + 4 \energyMem$ and for accumulate operations is $\energyAcc = \energyAdd + 3\energyMem$ \cite{sze2017efficient}.

For \emph{ANNs}, we obtain
\begin{align*}
\energyAnn=
\energyMac \width
(
\inputDim
+(\depth-1)\width
+\outputDim)+
(\energyAdd+\energyAct+5\energyMem)\depth\width + \energyAcc\outputDim\:.
\end{align*}
Absorbing constant energy coefficients and using $\width\geq 1$, we get
$\energyAnn
=
\calO(
\depth\width^2
+
(\inputDim+\outputDim)\width)$,
which highlights that the dominant cost comes from the matrix-vector multiplications in the hidden layers, with linear contributions from the input and output layers.
Due to the recurrent connections, the \emph{RNN} requires more operations and memory accesses than the ANN and has the energy demand
\begin{equation*}
\energyRnn = T \big[\energyAnn(\inputDim, \outputDim,\width, \depth,0,1) + \energyMac
\big(
\depth \width^2
\big) +
(2\energyAdd+6\energyMem)\depth\width \big]\:,
\end{equation*}
but it has the same per-timestep complexity as the ANN, 
$\energyRnn=
\mathcal{O}( T \cdot (
\depth\width^2
+
(\inputDim+\outputDim)\width))$.
The \emph{SNN}, in contrast, communicates through binary spikes and replaces the multiply-and-accumulate operations in the hidden layers by accumulate operations. Only the first layer still requires multiplications, since it encodes the real-valued input. This leads to the following estimated energy demand:
\begin{align*}
\energySnn = T [
\energyMac \inputDim \width +
\energyAcc \width ((\depth-1)\width + \outputDim)(1 - S) +
(3\energyAdd+\energyAct+9\energyMem+3\energyMemOne)\depth\width +
2\energyAcc\outputDim
]\:.
\end{align*}
Absorbing constant energy coefficients, we have
$
\energySnn=\calO(T (\width\inputDim+\width(1-S)(\depth\width + \outputDim)+\depth\width+\outputDim))
$,
showing the linear dependence on the sequence length and the sparsity.
\section{Energy-Expressivity Trade-Off}
\textbf{Expressivity Based on Upper Bounds for Regions.}
We evaluate the expressivity of the neural networks based on the partitioning of the input space generated by the hyperplane arrangement induced by the affine transformations of the linear layers.\footnote{This serves as a unified measure of expressivity, given that ANNs, RNNs, and SNNs represent fundamentally different classes of functions. It is suitable for classification tasks, as we assume constant and linear regions to be equally significant.}
The upper bound on the maximum number of distinct linear regions of the ANN is $\regionsANN = 2^{\depth\width}$, which corresponds to every hyperplane generated by a neuron bisecting all regions generated by the other neurons.
For RNNs, every hidden neuron is evaluated once per timestep, and each evaluation contributes one hyperplane, so $\regionsRNN = 2^{T\depth\width}$.
In SNNs, only the first layer partitions the continuous input space, while the subsequent layers operate on discrete spike representations \cite[Sec. 6]{nguyen2025time}. Therefore, SNNs cannot increase expressivity with depth, and deeper layers can only merge or preserve regions and cannot refine the input partitioning, implying that the maximum number of constant regions is fixed by the first layer and bounded by $\regionsSNN = 2^{T\width}$.
While a strictly fixed spike sparsity $\sparsity$ reduces the maximum achievable number of regions to $\binom{T\width}{(1-\sparsity)T\width}$, we treat $2^{T\width}$ as the structural capacity limit.

\textbf{Energy Comparison With Respect to Expressivity.}
To establish an architecture-independent energy comparison, we normalize the networks' energy by the logarithm of the upper bound on their
capacity.
We relate the ANN depth to the number of timesteps, $\depthANN = \alpha T$, fix the RNN depth to $\depthRNN = \beta$, and choose a single-layer SNN, $\depthSNN = 1$.
Hidden layer width~$\width$ and output dimension~$\outputDim$ are fixed. The input dimension is~$\inputDim$ for RNN and SNN as well as~$\inputDim T$ for the ANN. 
Under these assumptions, we get the following relation between the upper bounds on the regions: $\regionsSNN = 2^{(1-\alpha)T\width} \cdot  \regionsANN$ and $\regionsSNN = 2^{(1-\beta)T\width} \cdot \regionsRNN$.
We define the expressivity-normalized efficiency ratios $\eta_\text{ANN}(\alpha)$ and $\eta_\text{RNN}(\beta)$ as the energy consumption normalized by the logarithmic expressivity ($\log_2 \regions$):
\begin{align}
\label{eq:energy_ratio}
    \eta_\text{ANN}(\alpha) &:=  \frac{\alpha \ \energySnn(\inputDim, \outputDim, \width, 1, S,T)}{ \energyAnn(\inputDim T, \outputDim, \width, \alpha T, 0, 1)}\:,\qquad
    \eta_\text{RNN}(\beta) := \frac{ \beta \ \energySnn(\inputDim, \outputDim, \width, 1, S,T) }{\energyRnn(\inputDim, \outputDim, \width, \beta, 0, T)}\:.
\end{align}
An SNN is more efficient with respect to its expressivity if $\eta_\text{ANN}(\alpha)<1$ and $\eta_\text{RNN}(\beta) < 1$.
\begin{theorem}[Efficiency Ratio]
\label{thm:efficiency_ratio}
Consider an SNN with a single layer ($\depthSNN=1$), input dimension $\inputDim$, width $\width$, output dimension $\outputDim$, temporal sequence length $T$, and spike sparsity $S \in (0,1)$, compared against either an ANN with depth $\depthANN = \alpha T$ and input dimension $\inputDim T$, or an Elman RNN with depth $\depthRNN = \beta$ and input dimension $\inputDim$.
Each of the conditions $\eta_\text{ANN}(\alpha)<1$ and $\eta_\text{RNN}(\beta)<1$, jointly written as $\eta(\lambda)<1$, holds if and only if
    $
    A\width^2 + B\width + C > 0,
    $
    where
    $A = \energyMac(c\lambda-1)$,
    $B = \energyMac\outputDim + \energyMac\inputDim\tau(1-\lambda) - \lambda\energyAcc\tau\outputDim(1-S) + \lambda\delta$, and
    $C = \energyAcc\outputDim(1 - 2\lambda\tau)$,
    with $(\tau, c, \delta) = (T,\,T,\,-\energyDiff T)$ for the ANN and $(1,\,2,\,\energyDiffR)$ for the RNN, where $\energyDiff = 2\energyAdd + 4\energyMem + 3\energyMemOne$ and $\energyDiffR = 2\energyMem - 3\energyMemOne$.
\end{theorem}
\begin{figure*}
    \centering
    \includegraphics[width=1.0\linewidth]{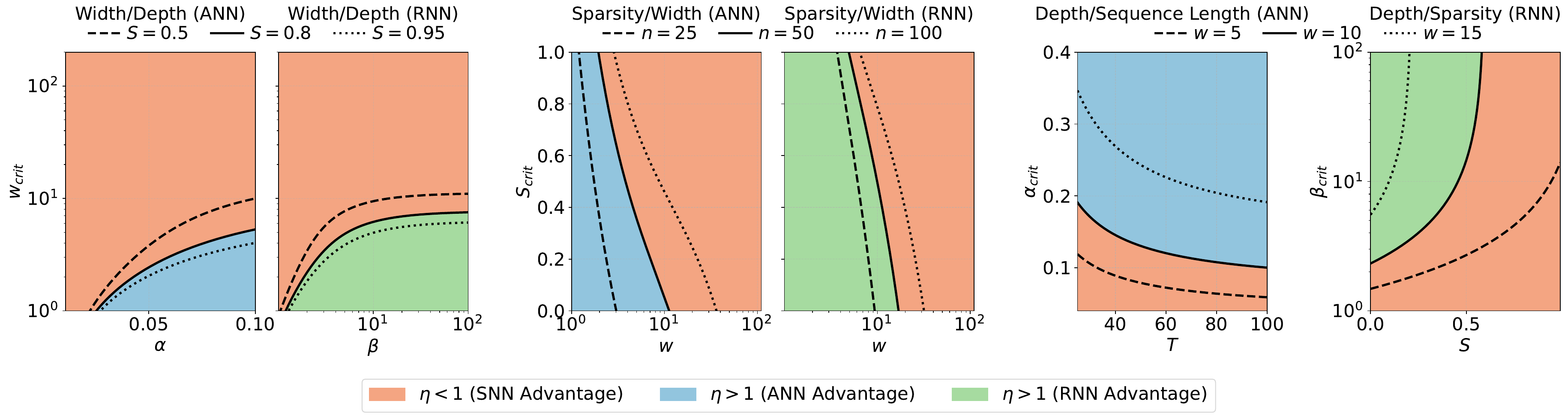}
    \caption{Evolution of efficiency thresholds. Critical values for width $\width_\text{crit}$, sparsity $\sparsity_\text{crit}$, and depth scaling factors $\alpha_\text{crit}$ and $\beta_\text{crit}$ under parameter shifts. Each group shows the ANN (left) and the RNN (right). Fixed base parameters: $\inputDim=1$, $\outputDim=50$, $\width=10$, $T=100$, $\sparsity=0.5$, $\alpha=0.05$ for the ANN, and $\beta=14$ for the RNN. Shading refers to the solid curve. Although $\depthANN=\alpha T\geq 1$ must be an integer, we treat $\alpha$ as a continuous relaxation.}
    \label{fig:energy_expressivity}
\end{figure*}

Below, we collect the thresholds for the width $\width$, the sparsity $\sparsity$, and the depth scaling parameter $\lambda$. Once these are crossed, the SNN is more efficient with respect to its capacity than the competitor.

\begin{remark}[Critical Thresholds]
\label{rem:thresholds}
For the width $\width$, suppose $A > 0$, which holds for the ANN if $\alpha T > 1$ and for the RNN since $\beta \ge 1$. The quadratic term in the inequality in \cref{thm:efficiency_ratio} then has exactly one positive root, $\width_{\text{crit}} = (-B + \sqrt{B^2 - 4AC})/(2A)$, and $\eta(\lambda) < 1$ holds precisely for $\width > \width_{\text{crit}}$. For a single-layer ANN ($A = 0$), the inequality is linear and holds for $\width > -C/B$, provided $B > 0$.
For the sparsity $\sparsity$, regardless of the sign of $A$, $\eta(\lambda) < 1$ holds exactly when $\sparsity > \sparsity_{\text{crit}}$ with $\sparsity_{\text{crit}} = 1 - (A\width^2 + \widehat{B}\width + C) / (\lambda\energyAcc\tau\outputDim\width)$, where $\widehat{B} := B + \lambda\energyAcc\tau\outputDim(1-\sparsity)$. The SNN is thus always more efficient if $\sparsity_{\text{crit}} \le 0$ and never if $\sparsity_{\text{crit}} \ge 1$.
For the depth scaling factor $\lambda$, the inequality is affine in $\lambda$ and yields the threshold
\begin{equation*}
    \lambda_{\text{crit}} = \frac{\energyMac\width^2 - \energyMac(\inputDim\tau + \outputDim)\width - \energyAcc\outputDim}{c\,\energyMac\width^2 - \energyMac\inputDim\tau\width - \energyAcc\tau\outputDim\width(1-\sparsity) + \delta\width - 2\tau\energyAcc\outputDim} \:.
\end{equation*}
If this denominator is positive, $\eta(\lambda) < 1$ holds for $\lambda > \lambda_{\text{crit}}$. If it is negative, the inequality reverses, and $\eta(\lambda) < 1$ holds for $\lambda < \lambda_{\text{crit}}$.
\end{remark}
The three thresholds exist for both competitors, but they are governed by different quantities. For the ANN, all of them depend on $T$. At fixed ANN depth $\depthANN$, a longer sequence favors the SNN. At fixed $\alpha$, as it is the case in \cref{fig:energy_expressivity}, the ANN depth grows with $T$, while $\alpha_\text{crit}$ decreases with $T$, and a longer sequence favors the ANN.
For the RNN, $\tau = 1$ removes $T$ from all three thresholds, which is why no sequence-length threshold exists in that comparison. \cref{fig:energy_expressivity} shows a selection of the resulting thresholds, and \cref{fig:energy_expressivity_all} in appendix \ref{app:energy_expressivity} shows the full set.
\section{Discussion}
\begin{wrapfigure}{r}{0.35\textwidth}
    \includegraphics[width=\linewidth]{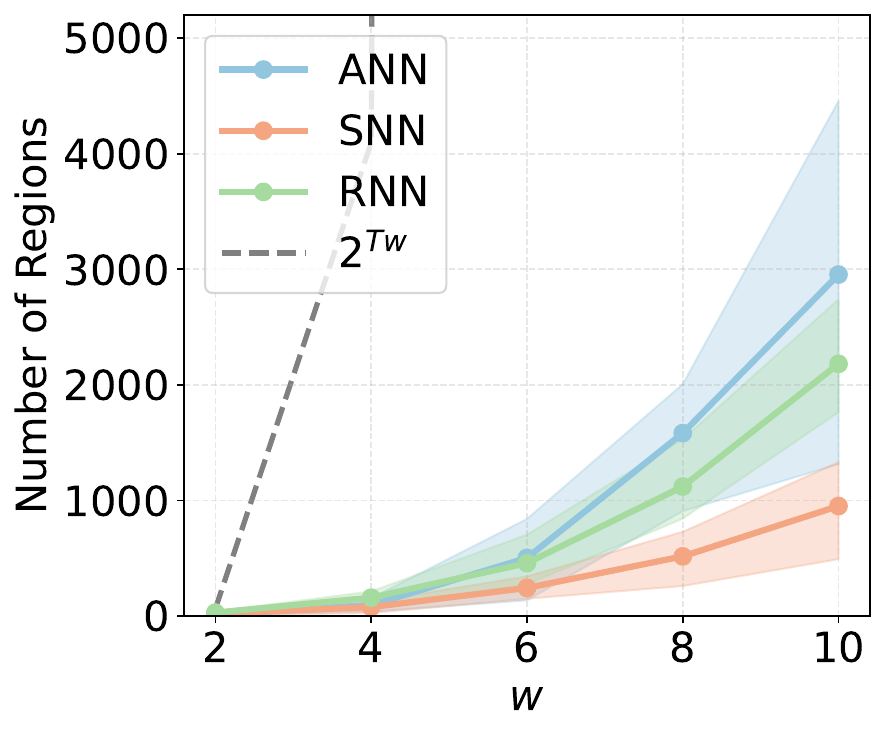}
    \caption{Evolution of the actual number of regions for increasing width. Mean and range over $10$ random seeds. Fixed base parameters:
$\inputDim=1$, $T=3$, $\alpha=\beta=1$.}
    \label{fig:regions_counted}
\end{wrapfigure}
Our comparison relies on general upper bounds on representational capacity, which may not be attained in practice. Tighter bounds are available for ANNs \cite{montufar2014number, serra2018bounding} and for SNNs with static inputs \cite{nguyen2025time} or time-to-first-spike encoding \cite{singh2023expressivity}. For RNNs, no comparable region bound is known: unrolling an Elman RNN over $T$ timesteps yields a weight-tied feedforward network of depth $TL$, to which the bounds of \cite{montufar2014number, serra2018bounding} apply. Deriving comparable bounds for time-series data remains an important direction for future work. 
Although the theoretical upper bounds for the architectures coincide at $\alpha=\beta=1$, \cref{fig:regions_counted} displays an empirical contrast: the actual number of regions in the ANN scales considerably faster with increasing network width than in the SNN, while the RNN lies in between.
To bridge this discrepancy between upper bounds and empirical behavior, future theoretical analysis could move toward average-case bounds or high-probability guarantees.

\begin{ack}
MK and GK acknowledge support by the project ``Next Generation AI Computing (gAIn)” funded by the Bavarian Ministry of Science and the Arts (StMWK Bayern) and the Saxon Ministry for Science, Culture, and Tourism (SMWK Sachsen).
\end{ack}

{
\small
\bibliographystyle{plain}
\bibliography{references}
}
\appendix

\section{Technical Appendices and Supplementary Material}

\subsection{Neural Networks}
\label{app:neural_networks}
\begin{definition}[Artificial Neural Network {\cite[Sec. 5.1]{bishop2006}}]
    Let $x \in \bbR^{\inputDim}$ be an input. For each layer $\ell \in [L+1]$, the output $h_\ell \in \bbR^{n_\ell}$ and the output of the network $y = \Phi(x) \in \bbR^{\outputDim}$ are given by
    \begin{align*}
        h_0 &= x\:,\\
        h_\ell &= \phi_\ell( W_\ell h_{\ell-1} + b_\ell ), \text{ for $\ell \in [\depth]$}\:,\\
        y &= h_{\depth+1} = W_{\depth+1} h_{\depth} + b_{\depth+1}\:,
    \end{align*}
    with weight matrices $W_\ell \in \bbR^{n_\ell \times n_{\ell-1}}$, biases $b_\ell \in \bbR^{n_\ell}$, and element-wise activation functions $\phi_\ell$.
\end{definition}
\begin{definition}[Multi-Layer Elman Recurrent Neural Network \cite{elman1990finding}]
Let $(x(t))_{t\in [T]}\in \bbR^{\inputDim\times T}$ be an input sequence. For each layer $\ell \in [\depth+1]$, the output $h_\ell \in \bbR^{n_\ell}$ and the output of the network $y = \Phi(x) \in \bbR^{\outputDim}$ are given by
    \begin{align*}
        h_0(t) &= x(t)\:,\\
        h_\ell(t) &= \phi_\ell(W^i_\ell h_{\ell-1}(t) + b^i_\ell + W^h_\ell h_{\ell}(t-1) + b^h_\ell )\:,\\
        y(t) &= h_{L+1}(t) = W_{L+1} h_L(t) + b_{L+1}\:,
    \end{align*}
with weight matrices $W^i_\ell\in \bbR^{n_\ell \times n_{\ell-1}}, W^h_\ell \in \bbR^{n_\ell \times n_{\ell}}$, biases $b^i_\ell, b^h_\ell \in \bbR^{n_\ell}$, and element-wise activation functions $\phi_\ell$. Here, $W^i_\ell$ and $b^i_\ell$ denote the parameters for the inputs of the layer, and $W^h_\ell$ and $b^h_\ell$ describe the parameters for the hidden inputs, i.e., the output of the layer in the previous timestep.
\end{definition}
\begin{definition}[Leaky Integrate-and-Fire (LIF) Spiking Neural Network {\cite[Sec. 2]{nguyen2025time}}]
\label{def:lif-snn}
    Let $(x(t))_{t\in [T]}\in \bbR^{\inputDim\times T}$ be an input sequence. For each layer $\ell \in [\depth+1]$, the output spikes $s_\ell(t) \in \{0,1\}^{n_\ell}$, the membrane potential $u_\ell(t) \in \bbR^{n_\ell}$ at timestep $t\in[T]$, and the output of the network $y = \Phi(x) \in \bbR^{\outputDim}$ are given by
    \begin{align*}
        s_0(t) &= x(t)\:,\\
        u_\ell(t)&=\gamma_\ell u_\ell (t-1)+W_\ell s_{\ell-1}(t)+b_\ell -\theta_\ell s_{\ell} (t-1)\:,\\
        s_\ell(t) &= H (u_\ell(t)-\theta_\ell \1_{n_\ell})\:,\\
        y &= D((u_{L+1}(t))_{t \in [T]})\:,
    \end{align*}
    with weight matrices $W_\ell \in \bbR^{n_\ell \times n_{\ell-1}}$ and biases $b_\ell \in \bbR^{n_\ell}$. The SNN-specific temporal parameters are the initial membrane potential $u_\ell(0)\in \bbR^{n_\ell}$, leaky terms $\gamma_\ell \in [0,1]$, and the thresholds $\theta_\ell \in (0,\infty)$. $H$ is the Heaviside function (applied entry-wise). The output encoding $D$ maps the time-series $(u_{\depth+1}(t))_{t\in [T]} \in \bbR^{\outputDim\times T}$ to an output. We choose $y = u_{L+1}(T)$.
\end{definition}
For simplicity, we focus on SNNs with integrate-and-fire (IF) neurons, where $\gamma = 1$.

\subsection{Energy Derivation}
\label{app:energy_derivation}
\begin{figure*}
    \centering
    \includegraphics{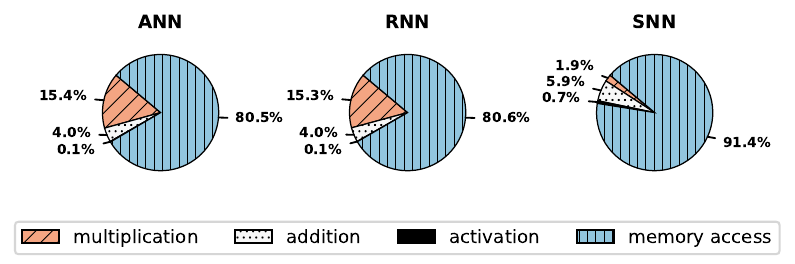}
    \caption{Distribution of total energy demand. Relative energy demand of arithmetic operations and memory accesses.}
    \label{fig:energy_comparison}
\end{figure*}

To estimate the energy demand of ANNs, RNNs, and SNNs, we quantify the total number of operations required for inference. 
This approach aligns with existing literature \cite{dampfhoffer2023are, lemaire2023analytical, yan2025reconsidering}.
\paragraph{Artificial Neural Networks.}
For a given layer $\ell$, the feedforward process requires $n_\ell \cdot n_{\ell -1}$ multiplications. 
Consequently, the total number of multiplications ($\annMult$) across the network is defined as
\begin{equation}
    \annMult = \underbrace{\inputDim \width}_\text{first hidden layer} + \underbrace{(\depth-1) \width^2}_\text{remaining hidden layers} + \underbrace{\outputDim \width}_\text{output layer}\:.
\end{equation}
The cumulative number of additions ($\annAdd$) accounts for both the accumulation of weighted inputs ($n_\ell \cdot n_{\ell -1}$ per layer) and the subsequent addition of bias terms ($n_\ell$ per layer), yielding
\begin{equation}
    \annAdd = \annMult+\underbrace{\depth \width + \outputDim}_\text{bias}\:.
\end{equation}
Finally, excluding the output layer, the activation function is applied to each neuron, resulting in a total number of activations ($\annAct$) given by
\begin{equation}
    \annAct = \depth \width\:.
\end{equation}
In standard ANNs, each multiply-and-accumulate (MAC) operation requires three read operations for input, weight, and partial sum, and one write operation to store the updated partial sum \cite{sze2017efficient}. Beyond these primary operations, the system must account for bias loading and saving as well as activation processing. The total number of read and write operations is defined as
\begin{equation*}
    \annMem =  \underbrace{3 \cdot \annMult}_\text{inputs, weights, partial sum} + \underbrace{2  (\depth \width + \outputDim)}_\text{biases, partial sum} + \underbrace{\depth \width}_\text{activation}+ \underbrace{\annMult + 2 \depth \width + \outputDim}_\text{write updated partial sum}\:.
\end{equation*}
\paragraph{Recurrent Neural Networks.}
The total number of multiplications ($\rnnMult$) in the RNN consists of the number of multiplications required by an ANN and additional multiplications for the recurrent layer-layer connections
\begin{equation}
    \rnnMult = \underbrace{\inputDim \width}_\text{first hidden layer} + \underbrace{(\depth-1) \width^2}_\text{remaining hidden layers} + \underbrace{\outputDim \width}_\text{output layer} +
    \underbrace{\depth \width^2}_\text{layer-layer connection}\:.
\end{equation}
The same applies to the total number of additions ($\rnnAdd$), which includes, along with the ANN additions, the additions for the recurrent connections as well as the combination of the results from the input and recurrent connections
\begin{equation}
    \rnnAdd = \rnnMult+\underbrace{\depth \width + \outputDim}_\text{bias} + \underbrace{\depth \width}_\text{layer-layer bias} +\underbrace{\depth \width}_\text{combine}\:.
\end{equation}
The number of activations ($\rnnAct$) is defined as
\begin{equation}
    \rnnAct = \depth \width\:.
\end{equation}
For the number of memory accesses ($\rnnMem$), additional read and write operations for the bias and partial sum of the recurrent connections and for the combinations are required and can be computed as
\begin{align*}
    \rnnMem &=  \underbrace{3 \rnnMult}_\text{inputs, weights, partial sum} + \underbrace{4 \depth \width + 2\outputDim }_\text{biases, partial sum} + \underbrace{2 \depth \width}_\text{combination} + \underbrace{\depth \width}_\text{activation} + \underbrace{\rnnMult + 4\depth \width+ \outputDim}_\text{write updated partial sum}\:.
\end{align*}
\paragraph{Spiking Neural Networks.}
Since the exact number of operations depends on the underlying hardware architecture, we introduce a set of simplifying assumptions to quantify the computational requirements of SNNs.
We assume that all input spikes are processed concurrently at each timestep. The computation is analogous to a linear layer in an ANN, where non-spiking neurons -- and thus their associated weights -- are omitted. 
Let $S_{\ell}(t)$ denote the sparsity of the output of layer $\ell$ at time $t$:
\begin{equation}
    S_{\ell}(t) = \frac{1}{n_\ell}\left(n_\ell - \sum_{i\in [n_\ell]}(s_\ell(t))_i\right)\:.
\end{equation}
Since the first layer of the SNN serves as an input encoding layer, the total number of multiplications ($\snnMult$) comprises all the multiplications of inputs and weights in the first layer:
\begin{equation*}
    \snnMult = \inputDim \width\:.
\end{equation*}
For the output encoding $D$, we choose to accumulate the membrane potential over all timesteps without emitting spikes. Hence, there is no reset in the output layer.
The total number of additions in layer $\ell$ comprises the accumulation of weights from spiking inputs, denoted as $n_{\ell-1} \cdot (1-S_{\ell-1}(t)) \cdot n_\ell$, the addition of the bias terms $n_\ell$ as well as the membrane potential update $n_\ell$, and the potential membrane reset $n_\ell$ for all layers excluding the output layer.
The total number of additions ($\snnAdd$) is thus defined as
\begin{align}
    \snnAdd = \underbrace{\inputDim \width}_\text{first hidden layer}+\underbrace{((\depth-1)\width+\outputDim)\cdot \width\cdot (1-\sparsity)}_\text{remaining hidden and output layers}+\underbrace{2 \depth \width +\outputDim}_\text{membrane} +\underbrace{\depth \width+\outputDim}_\text{bias}\:,
\end{align}
where $S$ is the overall sparsity over all timesteps and all layers except the output layer.
The number of activations ($\snnAct$), corresponding to the calculation of $s_\ell$, the application of the Heaviside step function to the difference between the membrane potential and the threshold, is given by
\begin{equation*}
    \snnAct = \depth \width\:.
\end{equation*}
For SNNs, we distinguish between standard multi-bit memory accesses ($\snnMem$) and single-bit operations ($\snnMemOne$) representing spike data. 
For each layer, the process involves reading the input spikes, synaptic weights, biases, current membrane potentials, and the output spikes from the previous timestep.
In the first layer, this comprises the external network input.
The corresponding write operations consist of the updated membrane potential and the output spikes $s_\ell(t)$.
The total memory access operations are
\begin{align*}
    \snnMem &= \underbrace{3\inputDim\width}_\text{first layer} +\underbrace{2 (((\depth-1)\width + \outputDim)\cdot \width \cdot (1-\sparsity))}_\text{weights, partial sum} + \underbrace{2(\depth \width + \outputDim)}_\text{biases, partial sum} 
    + \underbrace{ (3\depth \width+2\outputDim)}_\text{mem. pot., partial sum} \\
    &\quad + \underbrace{\depth \width}_\text{mem. pot. for activation}+\underbrace{\inputDim \width + ((\depth-1) \width + \outputDim)\cdot \width \cdot (1-\sparsity) +\depth \width+ \outputDim}_\text{updated partial sum}+ \underbrace{2\depth \width+\outputDim}_\text{updated mem. pot.},\\
    \snnMemOne &= \underbrace{\depth \width}_\text{input spikes} + \underbrace{\depth \width}_\text{prev. output spikes} + \underbrace{\depth \width}_\text{curr. output spikes}\:.
\end{align*}
The share of each operation for a fixed network and data size ($\inputDim=2$, $\outputDim=2$, $\width = 32$, $\depth = 3$, $T = 10$, $\sparsity = 0.8$) in the energy demand is depicted in \cref{fig:energy_comparison}, analytically determined by counting the operations in the respective network.
\subsection{Energy-Expressivity Trade-Off}
\label{app:energy_expressivity}
\paragraph{Proof of \cref{thm:efficiency_ratio}.}
From the definition of the expressivity-normalized efficiency ratio \eqref{eq:energy_ratio}, we have
\begin{align}
    &\eta(\alpha)= \alpha \frac{\energySnn(\inputDim,\outputDim, \width, 1, \sparsity, T)}{\energyAnn(\inputDim T, \outputDim, \width, \alpha T, 0, 1)}< 1 \\
    \label{eq:ann-inequality}
    &\quad\iff \energyAnn(\inputDim T, \outputDim, \width, \alpha T, 0, 1) - \alpha \energySnn(\inputDim,\outputDim, \width, 1, \sparsity, T) >0\:.
\end{align}
Using the energy formulas and choosing the depth as $\depthANN = \alpha T$, $\depthSNN = 1$, and the input dimension as $\inputDimANN=\inputDimSNN T$, we obtain
\begin{align*}
   \energyAnn
   &= \energyMac \cdot
\Bigl(
\inputDim T\width
+(\alpha T-1)\width^2
+\outputDim\width
\Bigr)+
(\energyAdd+\energyAct+5\energyMem)\alpha T\width + \energyAcc \outputDim\:, \\
\alpha \energySnn
&= \alpha T \Bigl[
\energyMac\inputDim\width
+
\energyAcc\outputDim\width(1 - S)
+
(3\energyAdd+\energyAct+9\energyMem+3\energyMemOne)\width+
2\energyAcc\outputDim
\Bigr]\:.
\end{align*}
Substituting into the inequality \eqref{eq:ann-inequality} yields
\begin{equation}
\label{eq:inequality_ann}
\begin{split}
&\underbrace{\energyMac(\alpha T-1)}_{A(\alpha)}\width^2
+
\underbrace{\Big[
\energyMac \outputDim
+
\energyMac\inputDim T(1-\alpha)
- \alpha \energyAcc T \outputDim ( 1-\sparsity)
- \alpha T \energyDiff
\Big]}_{B(\alpha)} \width\\
& \quad +
\underbrace{\Big[
\energyAcc\outputDim(1 - 2\alpha T)
\Big]}_{C(\alpha)}> 0
\end{split}
\end{equation}
with  $\energyDiff= (2\energyAdd + 4\energyMem + 3\energyMemOne)$.
For the RNN, we have
\begin{align}
    &\eta(\beta)= \beta \frac{\energySnn(\inputDim,\outputDim, \width, 1, \sparsity, T)}{\energyRnn(\inputDim, \outputDim, \width, \beta, 0, T)}< 1 \\
    \label{eq:inequality-rnn}
    &\quad\iff \energyRnn(\inputDim, \outputDim, \width, \beta , 0, T) - \beta \energySnn(\inputDim,\outputDim, \width, 1, \sparsity, T) >0\:.
\end{align}
The argument is analogous to that for the ANN, with $\depthRNN = \beta$, $\inputDimRNN = \inputDimSNN$, and $\lambda = \beta$. With this parameter choice, the energy demand of the RNN is 
\begin{align*}
   \energyRnn
   &= T\Bigl[\energyMac
\Bigl(
\inputDim\width + \outputDim\width + (2\beta-1)\width^2
\Bigr)+
(3\energyAdd+\energyAct+11\energyMem)\beta\width + \energyAcc \outputDim\Bigr]\:.
\end{align*}
Therefore, both sides of the inequality \eqref{eq:inequality-rnn} carry the factor $T$, which cancels. Collecting powers of $\width$ gives
\begin{equation}
\label{eq:inequality_rnn}
\begin{split}
&\underbrace{\energyMac(2\beta-1)}_{A(\beta)}\width^2
+
\underbrace{\Big[
\energyMac \outputDim
+
\energyMac\inputDim(1-\beta)
- \beta \energyAcc \outputDim ( 1-\sparsity)
+ \beta \energyDiffR
\Big]}_{B(\beta)} \width\\
& \quad +
\underbrace{\Big[
\energyAcc\outputDim(1 - 2\beta)
\Big]}_{C(\beta)}> 0
\end{split}
\end{equation}
with $\energyDiffR = (2\energyMem - 3\energyMemOne)$.

Both inequalities \eqref{eq:inequality_ann} and \eqref{eq:inequality_rnn} have the form
\begin{equation}
\label{eq:inequality}
    A\width^2 + B\width + C > 0
\end{equation}
with
\begin{align*}
A &= \energyMac(c\lambda-1)\:, \\
B &= \energyMac \outputDim + \energyMac\inputDim\tau(1-\lambda) - \lambda \energyAcc \tau \outputDim (1-\sparsity) + \lambda \delta\:, \\
C &= \energyAcc\outputDim(1 - 2\lambda\tau)\:,
\end{align*}
where $(\tau, c, \delta) = (T,\,T,\,-\energyDiff T)$ for the ANN and $(1,\,2,\,\energyDiffR)$ for the RNN.
The factor $\tau$ describes the shape of the input sequence. For the ANN, which processes the concatenated input in a single pass, $\tau = T$, and for the RNN, which processes one timestep at a time like the SNN, $\tau = 1$.
The coefficient $c$ counts the hidden-to-hidden weight matrices per unit of $\lambda$, one per layer for the ANN and two for the RNN.
Analogously, $\delta$ is the difference between the competitor and the SNN in the remaining per-neuron costs per unit of $\lambda$. These include biases, activations, and state updates.

\begin{figure*}
    \centering
    \includegraphics[width=1.0\linewidth]{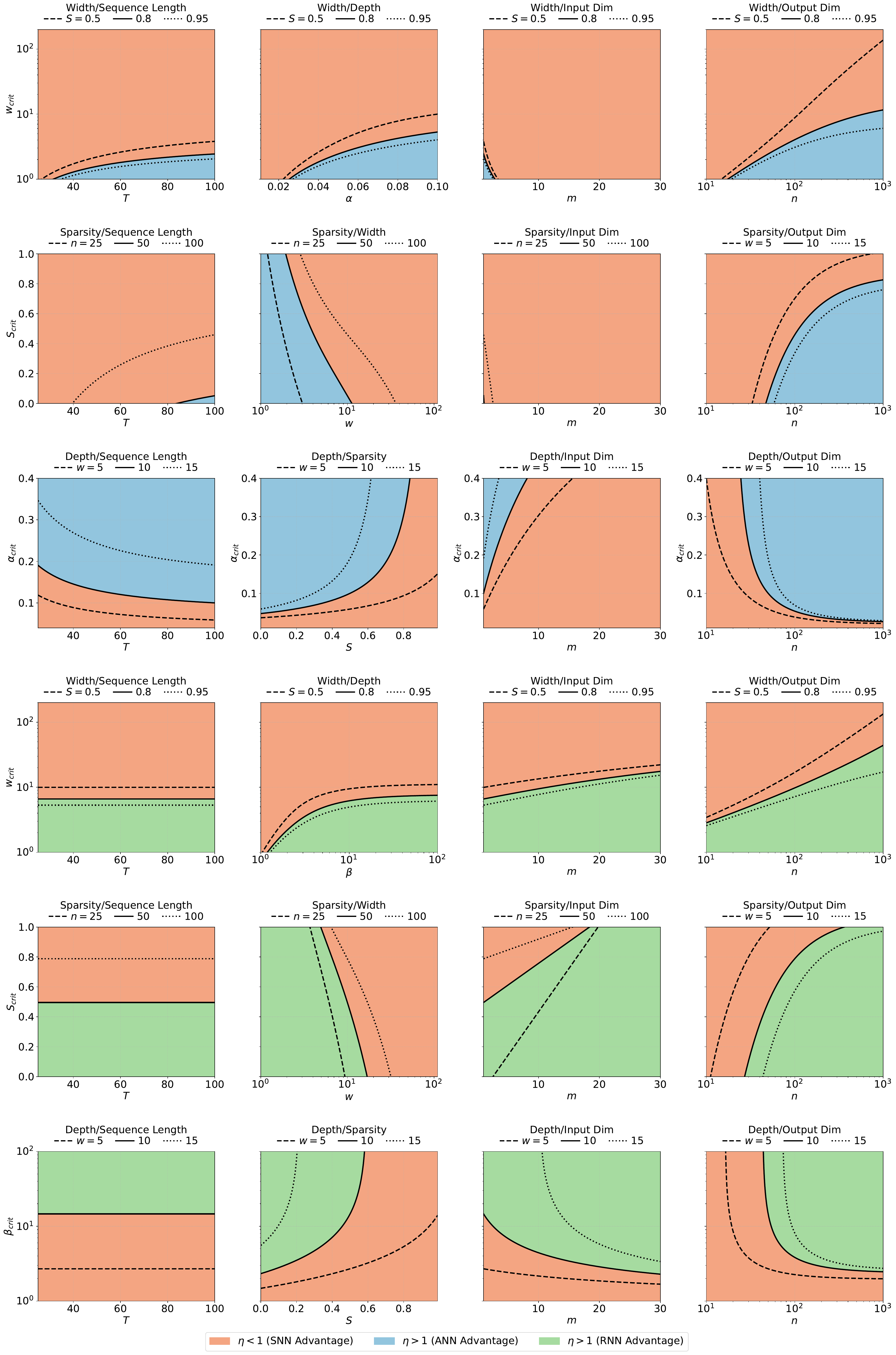}
    \caption{Full set of efficiency thresholds. Rows 1--3: ANN comparison. Rows 4--6: RNN comparison. Each row shows one threshold (width, sparsity, depth). Fixed base parameters: $\inputDim=1$, $\outputDim=50$, $\width=10$, $T=100$, $\sparsity=0.5$, $\alpha=0.05$ for the ANN, and $\beta=14$ for the RNN.}
    \label{fig:energy_expressivity_all}
\end{figure*}

The following paragraphs describe the derivations of the thresholds for this inequality, which are illustrated in \cref{fig:energy_expressivity_all}.

\paragraph{Critical Width Threshold $\width_\text{crit}$.}
For $A > 0$, the left-hand side of \eqref{eq:inequality} is a convex parabola in $\width$ with $C < 0$ whenever $\lambda\tau > 1/2$. It has exactly one positive root, and the inequality holds above
\begin{align*}
    \width_{\text{crit}} = \frac{-B + \sqrt{B^2 - 4AC}}{2A}\:.
\end{align*}
The condition $A > 0$ is equivalent to $c\lambda > 1$, which holds for the ANN if $\alpha T > 1$ and always for the RNN, since $\beta \ge 1$. It also implies $\lambda\tau > 1/2$, which is why $C < 0$ follows without further assumptions.
The case $A = 0$ only occurs for a single-layer ANN ($\alpha T = 1$). The left-hand side of \eqref{eq:inequality} is then linear in $\width$ with $C = -\energyAcc\outputDim < 0$. If $B > 0$, the inequality holds precisely for $\width > -C/B$, and otherwise for no width.

\paragraph{Critical Sparsity Threshold $\sparsity_\text{crit}$.}
Reordering inequality \eqref{eq:inequality} yields
\begin{equation*}
\lambda \energyAcc \tau \width \outputDim ( 1-\sparsity) <
\energyMac(c\lambda-1)\width^2
+
\Big[
\energyMac \outputDim
+
\energyMac\inputDim\tau(1-\lambda)
+ \lambda \delta
\Big] \width+
\energyAcc\outputDim(1 - 2\lambda\tau)\:.
\end{equation*}
Therefore, the SNN is superior if its sparsity fulfills
\begin{equation*}
\sparsity > 1-\frac{\energyMac(c\lambda-1)\width^2 + \big[\energyMac \outputDim + \energyMac\inputDim\tau(1-\lambda) + \lambda \delta\big] \width + \energyAcc\outputDim(1 - 2\lambda\tau)}{\lambda \energyAcc \tau \outputDim \width}\:.
\end{equation*}

\paragraph{Critical Depth Scaling Threshold $\lambda_\text{crit}$.}
The inequality \eqref{eq:inequality} is affine in $\lambda$. Collecting its terms gives
\begin{equation*}
\lambda \Big[c\,\energyMac\width^2
- \energyMac\inputDim\tau\width
- \energyAcc\tau\outputDim\width( 1-\sparsity)
+ \delta\width
- 2\tau\energyAcc\outputDim
\Big]
>
\energyMac \width^2
- \energyMac (\inputDim \tau + \outputDim)\width
- \energyAcc\outputDim \:,
\end{equation*}
hence the critical depth scaling factor threshold is
\begin{equation*}
    \lambda > \frac{\energyMac\width^2 - \energyMac(\inputDim\tau + \outputDim)\width - \energyAcc\outputDim}{c\,\energyMac\width^2 - \energyMac\inputDim\tau\width - \energyAcc\tau\outputDim\width(1 - \sparsity) + \delta\width - 2\tau\energyAcc\outputDim}\:,
\end{equation*}
where the denominator is the SNN’s savings per additional layer. If it is negative, the inequality reverses, and $\lambda_\text{crit}$ becomes an upper rather than a lower bound.

\subsection{Exact Number of Regions}
\label{app:exact_regions}
Computing the exact number of linear/constant regions constitutes a complex combinatorial problem.
Depending on the definition, it is already \#P- or NP-hard for ReLU networks with one hidden layer \cite{stargalla2025computational}.
As every neuron can have two states, active/inactive for ReLU and spike/no-spike for IF neurons,
the maximum number of regions is $\calK = 2^{\#\text{neuron evaluations}}$, i.e., $2^{L\width}$ for the ANN, $2^{TL\width}$ for the RNN, and $2^{T\width}$ for the SNN.
We use a depth-first search through the neurons to examine all possible combinations of neuron states leading to distinct regions in the input space, which we bound to $[-10,10]^{\inputDim T}$ for the ANN and $[-10,10]^{\inputDim \times T}$ for the SNN and the RNN.
For the ANN, the network is traversed layer-wise, while for the
SNN and the RNN, the algorithm operates timestep-wise.
Once the states of all earlier evaluations are fixed, the next pre-activation or membrane potential is affine in the input, including the recurrent term of the RNN and the reset by subtraction of the SNN, so each state adds a half-space.
At each node, a linear program checks whether the current cell contains a ball of radius $\varepsilon = 10^{-6}$, which discards empty and lower-dimensional cells.
The exact number of regions is the number of feasible leaf nodes of the search.
\cref{fig:regions_counted} reports the mean and range over 10 random seeds per width for $\inputDim = 1$, $T = 3$, and $\alpha = \beta = 1$.

\end{document}